\documentclass[runningheads]{llncs}
\usepackage{graphicx}
\usepackage{amsmath,amssymb} % define this before the line numbering.
\usepackage{color}
\usepackage{subfig}
\usepackage{cite}
\newcommand{\Lagr}{\mathcal{L}}
\DeclareMathOperator{\EX}{\mathbb{E}}% expected value
\begin{document}
\title{Unpaired Thermal to Visible Spectrum Transfer using Adversarial Training}
% Replace with your title

\titlerunning{Thermal to Visible Spectrum Transfer}
% Replace with a meaningful short version of your title
%
\author{Adam Nyberg\inst{1}\orcidID{0000-0001-8764-8499} \and
Abdelrahman Eldesokey \inst{1}\orcidID{0000-0003-3292-7153} \and
David Bergstr\"om \inst{2}\orcidID{0000-0003-2414-4482} \and
David Gustafsson\inst{2}\orcidID{0000-0002-4370-2286}}
%
%Please write out author names in full in the paper, i.e. full given and family names.
%If any authors have names that can be parsed into FirstName LastName in multiple ways, please include the correct parsing, in a comment to the volume editors:
%\index{Lastnames, Firstnames}
%(Do not uncomment it, because you may introduce extra index items if you do that, we will use scripts for introducing index entries...)
\authorrunning{A. Nyberg et al.}
% Replace with shorter version of the author list. If there are more authors than fits a line, please use A. Author et al.
%

\institute{Computer Vision Laboratory Link\"oping University, 581 83 Link\"oping, Sweden \and
Swedish Defence Research Agency (FOI), 583 30 Link\"oping, Sweden}
\maketitle              % typeset the header of the contribution
\begin{abstract}
Thermal Infrared (TIR) cameras are gaining popularity in many computer vision applications due to their ability to operate under low-light conditions. Images produced by TIR cameras are usually difficult for humans to perceive visually, which limits their usability. Several methods in the literature were proposed to address this problem by transforming TIR images into realistic visible spectrum (VIS) images. However, existing TIR-VIS datasets suffer from imperfect alignment between TIR-VIS image pairs which degrades the performance of supervised methods. We tackle this problem by learning this transformation using an unsupervised Generative Adversarial Network (GAN) which trains on unpaired TIR and VIS images. When trained and evaluated on KAIST-MS dataset, our proposed methods was shown to produce significantly more realistic and sharp VIS images than the existing state-of-the-art supervised methods. In addition, our proposed method was shown to generalize very well when evaluated on a new dataset of new environments.

\end{abstract}
\section{Introduction}

Recently, thermal infrared (TIR) cameras have become increasingly popular due to their long wavelength which allows them to work under low-light conditions. TIR cameras require no active illumination as they sense emitted heat from objects and map it to a visual heat map. This opens up for many applications such as object detection for driving in complete darkness and event detection in surveillance. In addition, the cost of TIR cameras have gone significantly down while their resolution have improved significantly, resulting in a boost of interest. However, one limitation of TIR cameras is their limited visual interpretability for humans which hinders some applications such as visual-aided driving.

\begin{figure}[tbp]
  \centering
  \subfloat[TIR image]{\includegraphics[width=0.33\textwidth]{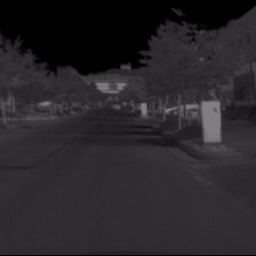}}
  \subfloat[Target VIS] {\includegraphics[width=0.33\textwidth]{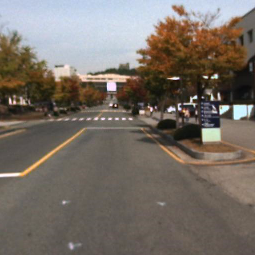}}
  \subfloat[Generated VIS] {\includegraphics[width=0.33\textwidth]{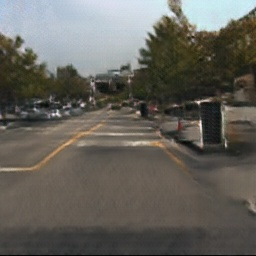}}
  \caption{An example of a TIR image (a), its corresponding VIS image (b) from the KAIST-MS dataset \cite{hwang2015multispectral} and the VIS image (c) generated by our proposed method using only the TIR image (a) as an input.}
  \label{fig:kaist-ex}
\end{figure}

To address this problem, TIR images can be transformed to visible spectrum (VIS) images which are easily interpreted by humans. Figure \ref{fig:kaist-ex} shows an example of a TIR image, the corresponding VIS image and the VIS image generated directly from the TIR image. This is similar to colorization problems, where grayscale VIS images are mapped to color VIS images. However, transforming TIR images to VIS images is inherently challenging as they are not correlated in the electromagnetic spectrum. For instance, two objects of the same material and temperature, but with different colors in the VIS image, could correspond to the same value in the TIR image. Consequently, utilizing all the available information, i.e. spectrum, shape and context, is very crucial when solving this task. This also requires the availability of enormous amount of data to learn the latent relations between the two spectrums.

In colorization problems, only the chrominance needs to be estimated as the luminance is already available from the input grayscale images. Contrarily, TIR to VIS transformation requires the estimation of both the luminance and the chrominance based on the semantics of the input TIR images. Besides, generating data for learning colorization models is easy as color images could be computationally transformed to grayscale images to create image pairs with perfect pixel-to-pixel correspondences.  In contrast, datasets containing registered TIR/ VIS image pairs are very few and requires a sophisticated acquisition systems for good pixel-to-pixel correspondence.

The KAIST Multispectral Pedestrian Detection Benchmark (KAIST-MS) \cite{hwang2015multispectral} introduced the first large-scale dataset with TIR-VIS image pairs. However, it was shown by \cite{Berg2018} that the TIR-VIS image pairs in KAIST-MS does not have a perfect pixel-to-pixel correspondence, with a pixel error of up to 16 pixels (5\%) in the horizontal direction. This would degrade the performance of supervised learning methods which tries to learn the pixel-to-pixel correspondences between image pairs and leads to corrupted output. To our knowledge there exist no large-scale public dataset of TIR-VIS image pairs with perfect pixel to pixel correspondence. Therefore, the method used for TIR to VIS image transformation need to control for this imperfection.

%%%%% Let's see first either we would include the new dataset or not
%In this paper we used the KAIST-MS dataset and a co-axial imaging system developed by Swedish Defence Research Agency (FOI) to capture a new dataset of TIR and VIS image pairs. The KAIST-MS dataset contain images of city and suburban environments while the FOI dataset contain natural environments such as forests and fields.

In this paper, we propose an unsupervised method for transforming TIR images, specifically long-wavelength infrared (LWIR), to visible spectrum (VIS) images. Our method is trained on unpaired images from the KAIST-MS dataset \cite{hwang2015multispectral} which allows it to handle the imperfect registration between the TIR-VIS image pairs. Qualitative analysis shows that our proposed unsupervised method produces sharp and perceptually realistic VIS images compared to the existing state-of-the-art supervised methods. In addition, our proposed method achieves comparable results to state-of-the-art supervised method in terms of L1 error despite being trained on unpaired images. Finally, our proposed method generalizes very well when evaluated on our new FOI dataset, which demonstrates the generalization capabilities of our method contrarily to the existing state-of-the-art methods.

\section{Related Work}
Colorizing grayscale images has been extensively investigated in the literature. Scribbles \cite{levin2004colorization} requires the user to manually apply strokes of color to different regions of a grayscale image and neighboring pixels with the same luminance should get the same color. Transfer techniques \cite{welsh2002transferring} use the color palette from a reference image and apply it to a grayscale image by matching luminance and texture. Both scribbles and transfer techniques require manual input from the user. Recently, colorization methods based on automatic transformation, i.e., the only input to the method is the grayscale image, have become popular. Promising results have been demonstrated in the area of automatic transformation using Convolutional Neural Networks (CNNs) \cite{cheng2015deep, IizukaSIGGRAPH2016, larsson2017colorization, zhang2016colorful} and Generative Adversarial Networks (GANs) \cite{pix2pix2016, cao2017unsupervised, suarez2017infrared} due to their abilities to model semantic representation in images.

In the infrared spectrum, less research has been done on transforming thermal images to VIS images. In \cite{limmer2016infrared}, a CNN-based method was proposed to transform near-infrared (NIR) images to VIS images. Their method was shown to perform well as the NIR and VIS images are highly correlated in the electromagnetic spectrum. Kniaz \textit{et al.} \cite{2017ISPAr52W4...41K} proposed VIS to TIR transformation using a CNN model as a way to generate synthetic TIR images. The KAIST-MS \cite{liu2016multispectral} dataset introduced the first realistic large-scale dataset of TIR-VIS image pairs which opened up for developing TIR-VIS transformation models. Berg  \textit{et al.} \cite{Berg2018} proposed a CNN-based model to transform TIR images to VIS images trained on the KAIST-MS dataset. However, the imperfect registration of the dataset caused the output from their method to be blurry and corrupted in some cases.

Generative Adversarial Networks (GANs) have shown promising results in unsupervised domain transfer \cite{pix2pix2016, zhu2017unpaired, zhu2017toward, liu2017unsupervised}. An unsupervised method does not require a paired dataset, hence, eliminating the need for pixel to pixel correspondence. In \cite{8014766, suarez2017learning}, GANs have demonstrated a very good performance on transferring NIR images to VIS images. Isola \textit{et al.} \cite{pix2pix2016} has shown some qualitative results from the KAIST-MS dataset as and example for domain transfer. Inspired by \cite{zhu2017unpaired}, we employ an unsupervised GAN to transform TIR images to VIS images which eliminates the deficiencies caused by the imperfect registration in the KAIST-MS dataset as the training set does not need to be paired. Different from \cite{Berg2018}, our proposed method produces a very realistic and sharp VIS images. In addition, our proposed method is able to generalize very well on unseen data from different environments.

\section{Method}
Here we describe our proposed approach for transforming TIR images to VIS image while handling data miss-alignment in the KAIST-MS dataset.

\begin{figure}[tbp]
  \centering
  \includegraphics[width=1\textwidth]{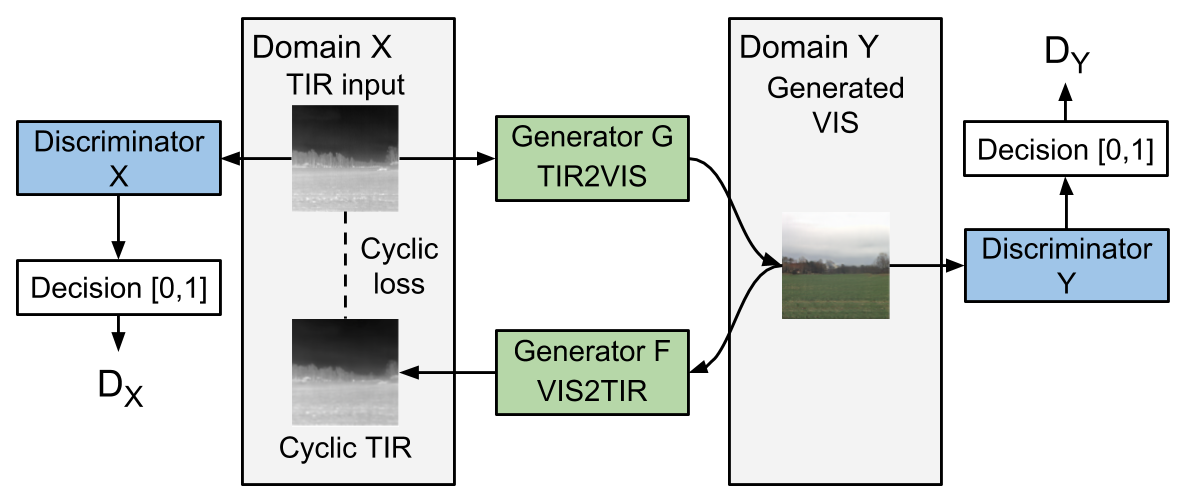}
  \caption{The unpaired model is mapping between two different domains, X (TIR) and Y (VIS), using $G : X \rightarrow Y$ and $F : Y \rightarrow X$. This model produces three different losses; two adversarial losses based on $D_Y$ and $D_X$, and one cycle consistency loss.}
  \label{fig:cycle_TIR2VIS}
\end{figure}

\subsection{Unpaired TIR-VIS Transfer}\label{sec:model-training}
Inspired by \cite{zhu2017unpaired}, we perform unsupervised domain transfer between TIR and VIS images. Given TIR domain $X$ with images $\{x_i:x_i \in X\}_{i=1}^{N}$ and VIS domain $Y$ with images $\{y_j:y_j \in Y\}_{j=1}^{M}$, we aim to learn two transformations $G$ and $F$ between the two domains as shown in Figure \ref{fig:cycle_TIR2VIS}. TIR input images are transformed from the thermal domain $X$ to the visible spectrum domain $Y$ using the generator $G$, while the generator $F$ performs in the opposite direction. An adversarial discriminator $D_X$ aims to discriminate between images $x$ and the transformed images $F(y)$, while another discriminator $D_Y$ discriminates between images $y$ and $G(x)$.

\subsection{Adversarial Training}
The main objective of a GAN with a cyclic loss is to learn the two transformations $G : X \rightarrow Y$, $F : Y \rightarrow X$ and their corresponding discriminators $D_X, D_Y$ \cite{zhu2017unpaired}. During training, the generator $G_{G}$ transforms an image $x \in X$ into a synthetic image $\hat{y}$. The synthetic image is then evaluated by the discriminator $D_{Y}$. The adversarial loss for the $G_{G}$ is defined as:
\begin{equation} \label{eq:generator_adversarial_loss}
\Lagr_{G_{G}}(G_{G}, D_Y, x) = \EX_{x\sim p_{data(x)}}[(D_{Y}(G_{G}(x)) - 1)^2]
\end{equation}
where $p_{data(x)}$ is data distribution in $X$. The loss value becomes large if the synthetic image $\hat{y}$ was able to fool the discriminator into outputting a value close or equal to one. On the other hand, the discriminator tries to maximize the probability for real images while also minimizing the output on synthetic images, achieved by minimizing the following formula:
\begin{equation} \label{eq:discriminator_adversarial_loss}
\Lagr_{D_{Y}}(D_Y, x, y) = \EX_{x\sim p_{data(x)}}[(D_{Y}(G_{G}(x)))^2] + \EX_{y\sim p_{data(y)}}[(D_{Y}(y) - 1)^2].
\end{equation}
The total adversarial loss for the \emph{G} transformation is then defined as:
\begin{equation} \label{eq:adversarial_loss}
\Lagr_{GAN}(G_{G}, D_{Y}, x, y) = \Lagr_{D_{Y}}(D_Y, y) + \min_{G_{G}} \max_{D_{Y}}\Lagr_{G_{G}}(G_{G}, D_Y, x)
\end{equation}
A similar loss is utilized to learn the transformation \emph{F}. To reduce the space of possible transformations, a cycle-consistency loss \cite{zhu2017unpaired} is employed which ensures that the learned transformation can map only a single input to the desired output. The cycle-consistency loss is defined as:
\begin{equation} \label{eq:cyclic_loss}
\begin{aligned}
    \Lagr_{cyc}(G_{G}, G_{F}) = \EX_{x\sim p_{data}(x)}[||G_{F}(G_{G}(x)) - x||_1] + \\
    \EX_{y\sim p_{data}(y)}[||G_{G}(G_{F}(y)) - y||_1]
\end{aligned}
\end{equation}
Combining the above losses gives our total loss which is defined as:
\begin{equation} \label{eq:cycleGAN_objective}
\begin{aligned}
    \Lagr(G_{G}, G_{F}, D_{X}, D_{Y}) = \Lagr_{GAN}(G_{G}, D_{Y}, x, y) + \\ \Lagr_{GAN}(G_{F}, D_{X}, y, x) + \\
    \lambda\Lagr_{cyc}(G_{G}, G_{F})
\end{aligned}
\end{equation}
where $\lambda$ is a factor used to control the impact of the cyclic loss.

\section{Experiments}
For evaluation, we compare our proposed method with the existing state-of-the-art method on TIR to VIS transfer TIR2Lab \cite{Berg2018}. The evaluation is performed on the KAIST-MS, the FOI dataset and the generalization capabilities of the evaluated methods are tested by training on the former and evaluating on the latter.

\subsection{Datasets}
\paragraph{The KIAST-MS dataset} \cite{hwang2015multispectral} contains paired TIR and VIS images which were collected by mounting a TIR and a VIS cameras on a vehicle in city and suburban environments. The KAIST-MS dataset consists of $33,399$ training image pairs and $29,179$ test image pairs captured during daylight with a spatial resolution of ($640\times512$). Because the TIR camera used, FLIR A35, is only capable of capturing images with a resolution of ($320\times256$) we resized all images to ($320\times256$). The images were collected continuously during driving, resulting in multiple image pairs being very similar. To remove redundant images, only every fourth image pair was included in the training set, resulting in $8,349$ image pairs. For the evaluation, all image pairs from the test set were used.

\paragraph{The FOI dataset} was captured using a co-axial imaging system capable of, theoretically, capturing TIR and VIS images with the same optical axis. Two cameras were used for all data collection, a TIR camera FLIR A65\footnote{http://www.flir.co.uk/automation/display/?id=56345} and VIS camera XIMEA MC023CG-SY\footnote{https://www.ximea.com/en/products/usb-31-gen-1-with-sony-cmos-xic/mc023cg-sy}. All images were cropped and re-sized to $320\times256$ pixels. This system was used to capture TIR-VIS image pairs in natural environments with fields and forests with a training set of $5,736$ image pairs and $1,913$ image pairs for the test set. The average registration error for all image pairs were between 0.8 and 2.2 pixels.

\subsection{Experimental Setup}
Since we use the architecture from \cite{zhu2017unpaired}, we crop all training and test images from the center to the size ($256\times256$). TIR2Lab \cite{Berg2018} was trained and evaluated on the full resolution of the images. All experiments were performed on GeForce GTX 1080 Ti graphics card \footnote{https://www.nvidia.com/en-us/geforce/products/10series/geforce-gtx-1080-ti}.

\paragraph{KAIST-MS dataset experiments}
A pretrained model for TIR2Lab \cite{Berg2018} trained on the KAIST-MS dataset was provided by the authors. Our proposed model (TIRcGAN) was trained from scratch for 44 epochs using the same hyperparameters as \cite{zhu2017unpaired}. Those parameters were batch size = $1$, $\lambda=10$ and for the ADAM optimizer we used learning rate$=2e-4$, $\beta_1=0.5$, $\beta_2=0.999$ and $\epsilon=1e-8$.  Some examples from pix2pix model \cite{pix2pix2016} on the KAIST-MS dataset were provided by the authors and are discussed in the qualitative analysis.

\paragraph{FOI dataset experiments}
 When training the TIR2Lab model, we used the same hyperparameters as mentioned in \cite{Berg2018}, except that we had to train for 750 epochs before it converged. For our TIRcGAN model, we trained for 38 epochs, using the same parameters as in the \textit{KAIST-MS dataset experiments}.

\paragraph{Evaluation Metrics}
For the quantitative evaluation we use $L_1$, root-mean-square error (RMSE), peak signal-to-noise ratio (PSNR) and Structural Similarity (SSIM) calculated between the transformed TIR image and the target VIS image. All metrics were calculated in the RGB color space normalized between 0 and 1 with standard deviation denoted as $\pm$.

\begin{table}[tbp]
\centering
\begin{tabular}{| c | c | c | c | c | c | c | }
 \hline
   Model & Trained on & Evaluated on & $L_1$ & RMSE & PSNR & SSIM \\ [0.5ex]
 \hline\hline

  TIR2Lab & KAIST-MS & KAIST-MS & \textbf{0.13}  & 0.46  & \textbf{14.7}  & {0.64}  \\
   & & & \textbf{$\pm$0.04} & $\pm$0.09 & \textbf{$\pm$2.20} & {$\pm$0.08}  \\ \hline
 TIRcGAN (Ours) & KAIST-MS & KAIST-MS & 0.15  & \textbf{0.21}  & 13.92  & 0.55 \\
    & & & $\pm$0.05 & \textbf{$\pm$0.06} & $\pm$2.48 & $\pm$0.10  \\
 \hline \hline

  TIR2Lab & FOI dataset & FOI dataset & 0.12 & 0.15 & 17.05 & 0.81  \\
    & & & $\pm$0.04 & $\pm$0.05 & $\pm$2.52 & $\pm$0.07  \\ \hline
   TIRcGAN (Ours) & FOI dataset & FOI dataset & \textbf{0.11}  & \textbf{0.14}  & \textbf{17.63}  & 0.77 \\
    & & & \textbf{$\pm$0.05} & \textbf{$\pm$0.05} & \textbf{$\pm$2.83} & $\pm$0.10  \\
 \hline \hline

 TIR2Lab & KAIST-MS & FOI dataset & 0.32 & 0.39 & 8.48 & 0.54  \\
    & & & $\pm$0.11 & $\pm$0.11 & $\pm$2.68 & $\pm$0.06  \\ \hline
  TIRcGAN (Ours) & KAIST-MS & FOI dataset & \textbf{0.20}  & \textbf{0.23}  & \textbf{13.10}  & {0.60} \\
    & & & \textbf{$\pm$0.07} & \textbf{$\pm$0.08} & \textbf{$\pm$2.82} & {$\pm$0.10}  \\
 \hline \hline

\end{tabular}
\caption{This table shows the qualitative results for the experiments calculated on RGB values normalized between 0 and 1. The standard deviation is denoted as $\pm$.}
\label{table:quant}
\end{table}

\subsection{Quantitative Results}
Table \ref{table:quant} shows the quantitative results for the state-of-the-art method TIR2Lab \cite{Berg2018} on the task of TIR to VIS domain transfer and our proposed method. Our method achieves comparable results to TIR2Lab in terms of $L_1$ despite the fact that our proposed method is unsupervised. On the other hand, our proposed method has a significantly lower RMSE than TIR2Lab which indicates its robustness against outliers. On the FOI dataset, our proposed method marginally outperforms TIR2Lab with respect to all evaluation metrics.

\paragraph{Model Generalization}
To evaluate generalization capabilities of methods in comparison, we train them on the KASIT-MS dataset and evaluate on the FOI dataset. The former was captured in city and urban environment, while the latter was captured in natural environments and forests. As shown in Table \ref{table:quant}, our proposed method maintains its performance to a big extent when evaluated on a different dataset. On the other hand, TIR2Lab model failed to generalize to unseen data.

\begin{figure}[t!]
  \centering
  \subfloat[TIR input]{\includegraphics[width=0.2\textwidth]{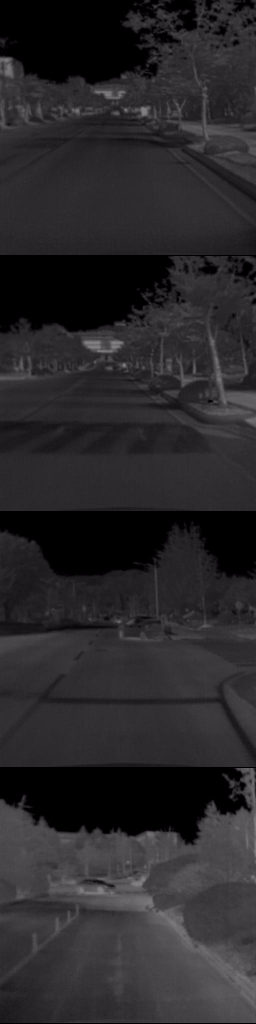}}
  \subfloat[pix2pix \cite{pix2pix2016}] {\includegraphics[width=0.2\textwidth]{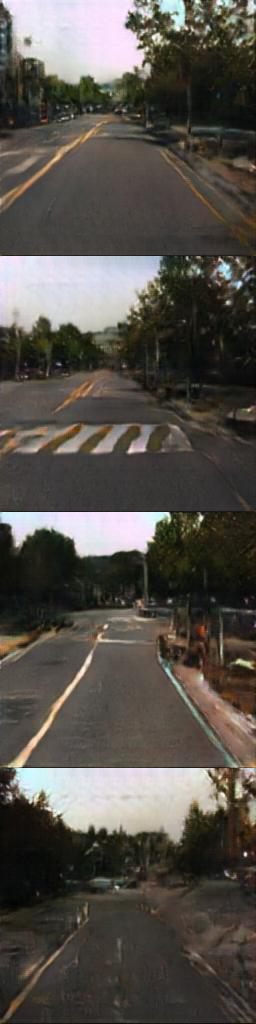}}
  \subfloat[TIR2Lab \cite{Berg2018}] {\includegraphics[width=0.2\textwidth]{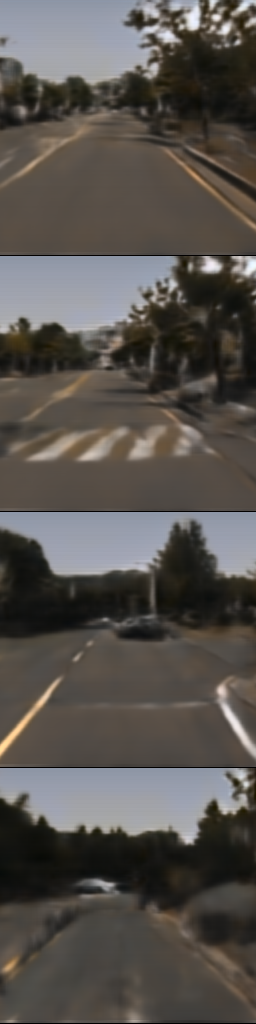}}
  \subfloat[TIRcGAN] {\includegraphics[width=0.2\textwidth]{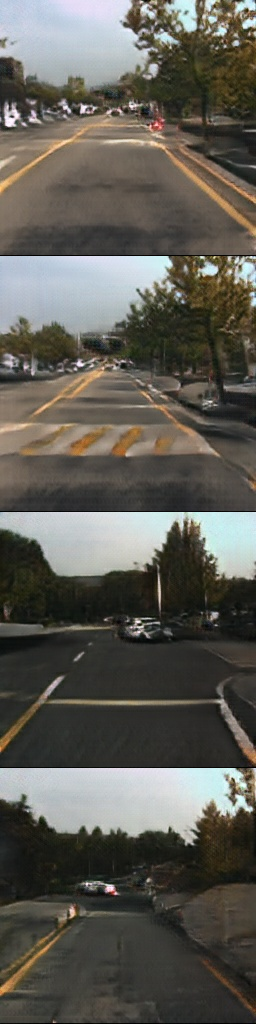}}
  \subfloat[Target VIS] {\includegraphics[width=0.2\textwidth]{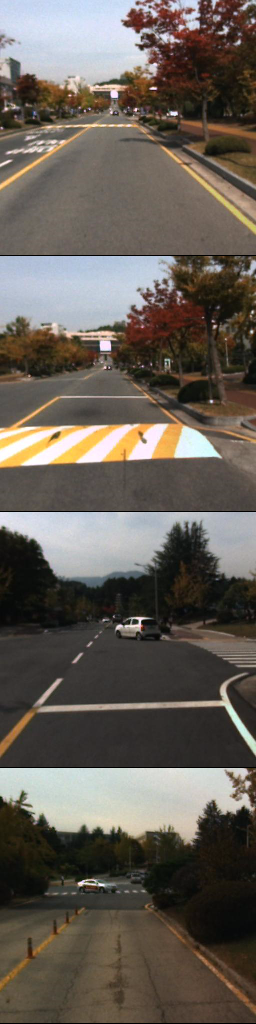}}
  \caption{Example images from the KAIST-MS dataset experiment. It is possible to note that the pix2pix and TIRcGAN produce sharper images than the TIR2Lab model. TIRcGAN model is able to distinguish between yellow and white lanes as seen in the third row.
  \emph{Note:} in the third row, the corresponding frame for the pix2pix model was not available, so the closest frame was used.}
  \label{fig:comp-kaist}
\end{figure}

\subsection{Qualitative Analysis}
\paragraph{The KAIST-MS dataset}
As shown in Figure \ref{fig:comp-kaist}, our proposed TIRcGAN  produces much sharper and saturated images than the TIR2Lab model on the KAIST-MS dataset. In addition, TIRcGAN is more inclined to generate smooth lines and other objects as an attempt to make the transformed TIR image look more realistic. On the other hand, images from TIR2Lab are quite blurry and lacks details in some occasions. Pix2pix performs reasonably and produces sharp images, however, objects and lines are smeared out in some cases.

\begin{figure}[t!]
  \centering
  \subfloat[TIR input]{\includegraphics[width=0.25\textwidth]{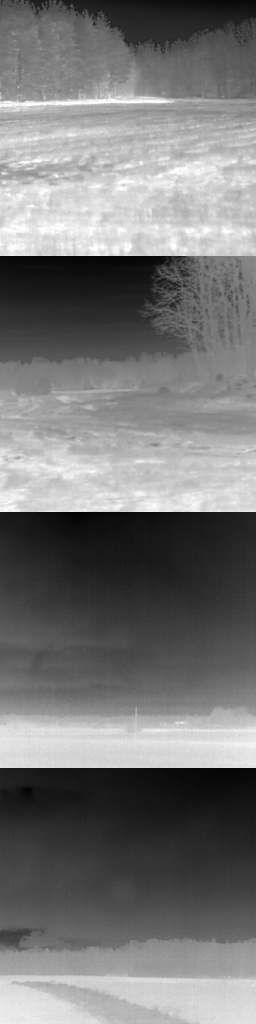}}
  \subfloat[TIR2Lab \cite{Berg2018}] {\includegraphics[width=0.25\textwidth]{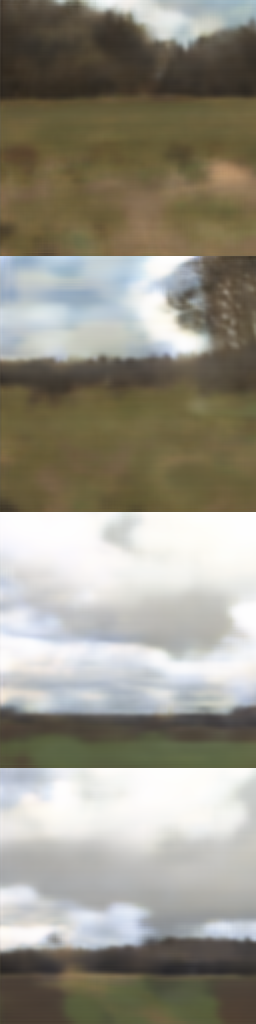}}
  \subfloat[TIRcGAN] {\includegraphics[width=0.25\textwidth]{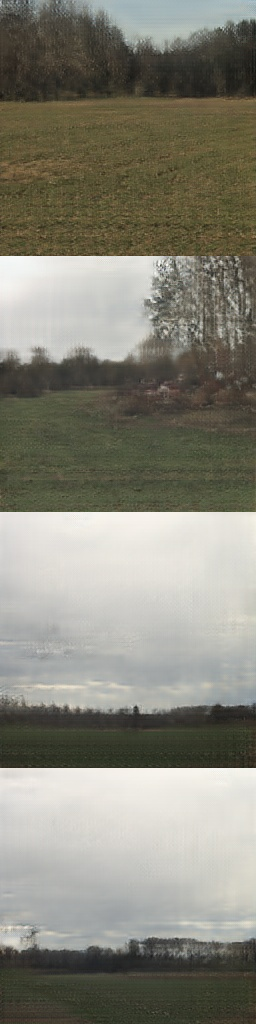}}
  \subfloat[Target VIS] {\includegraphics[width=0.25\textwidth]{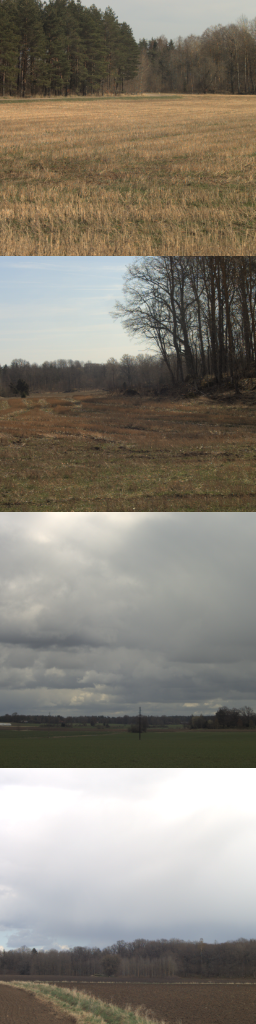}}
  \caption{Example images from the FOI dataset experiment. We can see that the unpaired model produce much sharper and more realistic images.}
  \label{fig:comp-our}
\end{figure}

\paragraph{The FOI dataset}
Figure \ref{fig:comp-our} show the results for the TIR2Lab and our proposed TIRcGAN models on the FOI dataset. TIRcGAN consistently outperform the TIR2Lab model when it comes to producing perceptually realistic VIS images. TIR2Lab produces blurry images that lacks a proper amount of details, while TIRcGAN produces a significant amount of details that are very similar to the target VIS image.

\begin{figure}[hb!]
  \centering
  \subfloat[TIR input]{\includegraphics[width=0.25\textwidth]{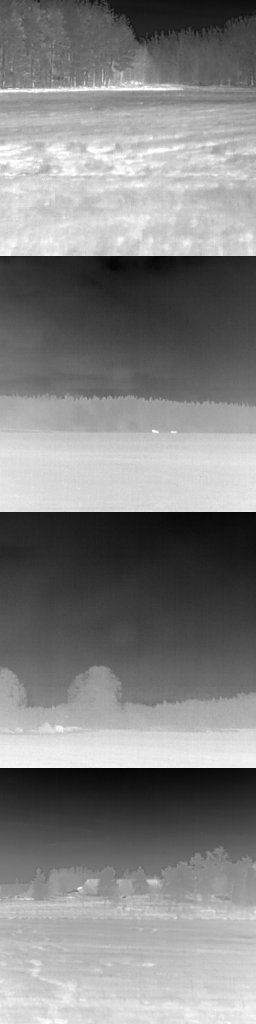}}
  \subfloat[TIR2Lab \cite{Berg2018}] {\includegraphics[width=0.25\textwidth]{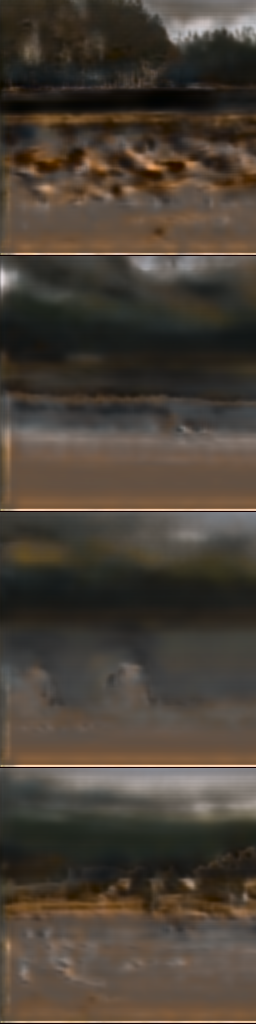}}
  \subfloat[TIRcGAN] {\includegraphics[width=0.25\textwidth]{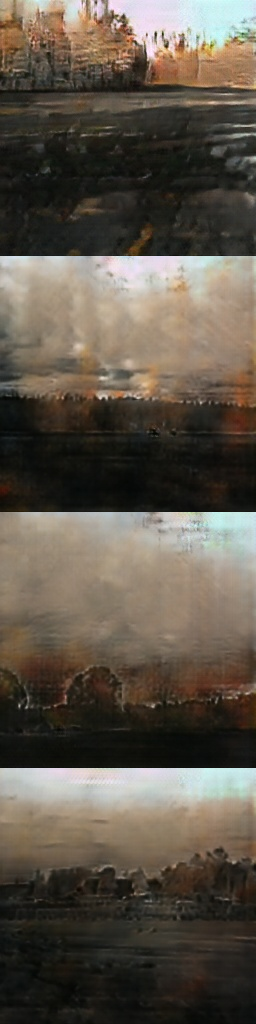}}
  \subfloat[Target VIS] {\includegraphics[width=0.25\textwidth]{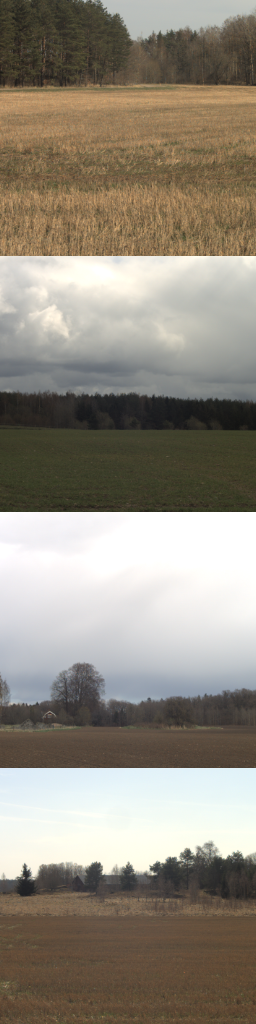}}
  \caption{Examples for methods output when trained on the KAIST-MS dataset \cite{hwang2015multispectral} and evaluated on the FOI dataset. Both models struggle with predicting the colors since the two datasets were captured in different environments. However, TIRcGAN can still predict the objects in the scene.}
  \label{fig:comp-gen}
\end{figure}

\paragraph{Model Generalization}
Figure \ref{fig:comp-gen} show the TIR2Lab and our TIRcGAN ability to generalize from one dataset collected on one environment to a new dataset from a different environment. Both models struggle at generating accurate colors or perceptually realistic images since the two datasets have different colors distribution. However, TIRcGAN was able to predict objects in the image with a reasonable amount of details contrarily to TIR2Lab which completely failed.

\subsection{Failure Cases}
\begin{figure}[ht!]
  \centering
  \subfloat[TIR input]{\includegraphics[width=0.2\textwidth]{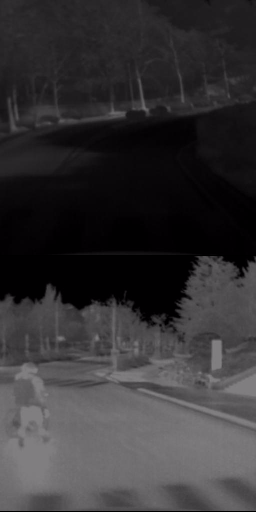}}
  \subfloat[pix2pix \cite{pix2pix2016}] {\includegraphics[width=0.2\textwidth]{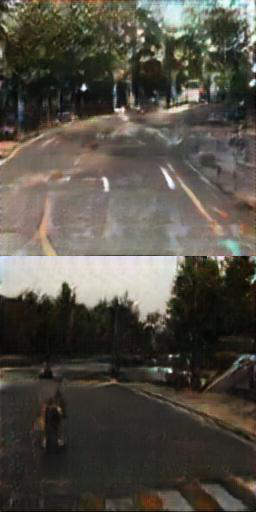}}
  \subfloat[TIR2Lab \cite{Berg2018}] {\includegraphics[width=0.2\textwidth]{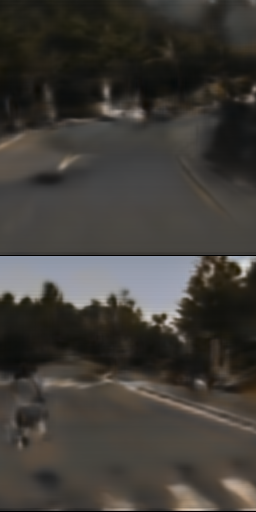}}
  \subfloat[TIRcGAN] {\includegraphics[width=0.2\textwidth]{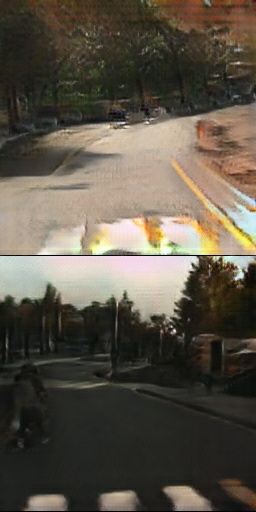}}
  \subfloat[Target VIS] {\includegraphics[width=0.2\textwidth]{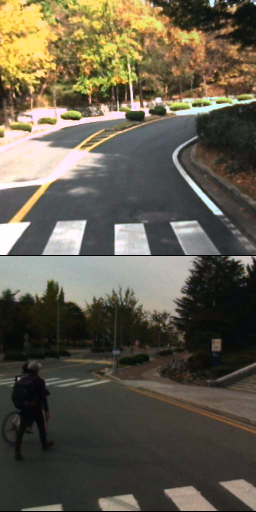}}
  \caption{Example images where different models fail on the KAIST-MS dataset. In the first row, we see that the model produces an inverse shadow, i.e., painting shadow only where there should not be shadow. In the second row we show that all the models struggle with producing perceptually realistic VIS images of humans.}
  \label{fig:comp-kaist-fail}
\end{figure}

\begin{figure}[ht!]
  \centering
  \subfloat[TIR input]{\includegraphics[width=0.25\textwidth]{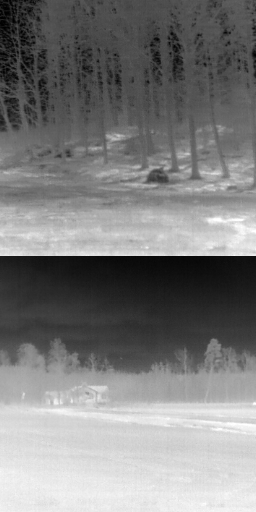}}
  \subfloat[TIR2Lab \cite{Berg2018}] {\includegraphics[width=0.25\textwidth]{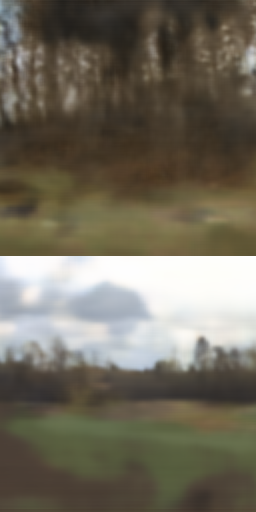}}
  \subfloat[TIRcGAN] {\includegraphics[width=0.25\textwidth]{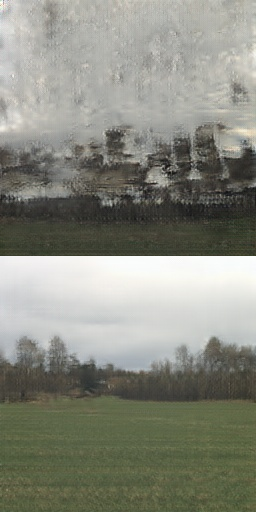}}
  \subfloat[Target VIS] {\includegraphics[width=0.25\textwidth]{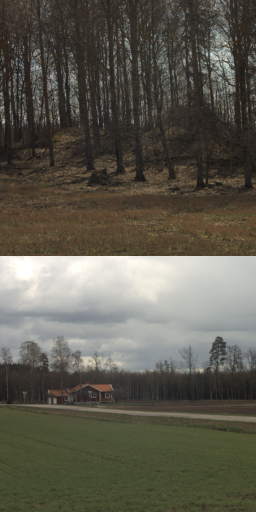}}
  \caption{Example images where models fail on the FOI dataset experiment. Here we see that the models are not able to accurately colorize houses and some roads.}
  \label{fig:comp-our-fail}
\end{figure}

Figure \ref{fig:comp-kaist-fail} shows examples where different models fail on the KAIST-MS dataset. For all methods, predicting humans is quite troublesome. Road crossing lines are also challenging as they are not always visible in the TIR images. Figure \ref{fig:comp-our-fail} shows some failure case on the FOI dataset. Both models fails in predicting dense forests, side-roads and houses since they are not very common in the dataset.

\section{Conclusions}
In this paper, we addressed the problem of TIR to VIS spectrum transfer by employing unsupervised GAN model that train on unpaired data. Our method was able to handle misalginments in the KAIST-MS dataset and produced perceptually realistic and sharp VIS images compared to the supervised state-of-the-art methods. When our method was trained on the KAIST-MS dataset and evaluated on the new FOI dataset, it maintained its performance to a big extent. This demonstrated the generalization capabilities of our proposed method.
%
% ---- Bibliography ----
%
% BibTeX users should specify bibliography style 'splncs04'.
% References will then be sorted and formatted in the correct style.
%
\bibliographystyle{splncs04}
\bibliography{mybibliography}
\end{document}